\documentclass{article}

\usepackage{microtype}
\usepackage{graphicx}
\usepackage{subcaption}
\usepackage{booktabs} 

\usepackage{hyperref}

\usepackage[preprint]{icml2026}

\usepackage{amsmath}
\usepackage{amssymb}
\usepackage{mathtools}
\usepackage{amsthm}

\usepackage[capitalize,noabbrev]{cleveref}
\usepackage{booktabs}   
\usepackage{colortbl}   
\usepackage{xcolor}     
\usepackage{multirow}   



\usepackage{algorithm}
\usepackage{algpseudocode}
\usepackage{float}
\usepackage{placeins}

\definecolor{rowblue}{RGB}{237, 246, 252}

\theoremstyle{plain}

\theoremstyle{definition}

\theoremstyle{remark}

\usepackage[textsize=tiny]{todonotes}

\icmltitlerunning{ScalingAttention: Intrinsic Sparse Attention Topology for vDiTs }

\begin{document}

\twocolumn[
  \icmltitle{ScalingAttention: Discovering Intrinsic Sparse Attention Topology for Video Diffusion Transformers}
\vspace{0.25em}
\begin{center}
{\large
\textbf{Ruiliang Zhou}$^{1,*,}$ \quad
\textbf{Xuecheng Wu}$^{2,*,\ddagger}$ \quad
\textbf{Kang He}$^{1}$ \quad
\textbf{Guangyun Han}$^{3}$ \quad
\textbf{Bin Liu}$^{3}$ \quad
\textbf{Qinqin Chen}$^{3}$ \quad
\textbf{Wende Xu}$^{4,\ddagger}$ \quad
\textbf{Qingjie Zhao}$^{2}$ \quad
\textbf{Chengru Song}$^{1}$
\par}

\vspace{0.45em}
{${}^1$ KlingAI Research \quad ${}^2$ Beijing Institute of Technology \quad ${}^3$ NVIDIA \quad ${}^4$ Tsinghua University \par}
\icmlcorrespondingauthor{}{zhouruiliang@kuaishou.com}

\end{center}


  \vskip 0.2in
]



\printAffiliationsAndNotice{$^*$ Equal contribution. $\ddagger$ Work done during internship at KlingAI Research.}

\begin{abstract}
While Diffusion Transformers (DiTs) have revolutionized high-fidelity video generation, their reliance on 3D full attention creates a quadratic computational bottleneck. Existing sparse methods face a dilemma: dynamic pruning suffers from prohibitive runtime overhead and memory fragmentation, while static heuristics fail to capture fine-grained dependencies. 
In this work, we propose ScalingAttention, a training-free framework grounded in a key inductive bias: while individual activations are input-dependent, the high-mass attention regions for each head rapidly converge to a stable, prompt-agnostic \textit{Intrinsic Sparse Topology}. This topology is weight-encoded, scale-invariant, and efficient to extract.
ScalingAttention decouples topology discovery from sparsity control via:
(1) \textbf{WEST} (Weight-Encoded Sparse Topology), which extracts a robust block-sparse prior mask offline to eliminate runtime search;
(2) \textbf{FAST} (Fidelity-Aware Sensitivity Tuning), which adaptively tunes head-wise sparsity based on diffusion fidelity requirements.
To ensure practical acceleration, we co-design a hardware-aligned bit-wise block-sparse kernel. Experiments on Wan2.1 show up to 1.90$\times$ end-to-end speedup with superior fidelity, establishing a new Pareto frontier over state-of-the-art baselines.
\end{abstract}

\section{Introduction}

Diffusion Transformers (DiTs) have recently achieved remarkable success in high-fidelity generative modeling, delivering state-of-the-art performance in both image and video synthesis \cite{wan2025wanopenadvancedlargescale,kong2025hunyuanvideosystematicframeworklarge}.
By combining diffusion-based iterative refinement with the expressive capacity of Transformers, DiTs effectively capture complex spatial structures and long-range temporal dependencies \cite{sd,dit,vit}.
However, these advances come at a substantial computational cost.
In video generation, DiTs rely heavily on 3D spatio-temporal self-attention, whose quadratic complexity with respect to the number of tokens leads to prohibitively expensive inference \cite{ptattention}.

For instance, generating a five-second video with 81 frames using the Wan2.1-14B model on an GPU requires nearly 30 minutes, with more than half of the end-to-end runtime dominated by 3D attention.

A natural approach to mitigating this bottleneck is sparse attention \cite{nsa,moba}.
Prior work has observed that attention distributions are inherently sparse, with only a small subset of critical tokens exerting a dominant influence on the output \cite{H20,attnsink}. 
Despite substantial recent efforts, existing sparse attention methods
for Video DiTs face a persistent dilemma.
Dynamic methods select important tokens at runtime, but incur substantial overhead and irregular execution that underutilize modern Tensor Cores \cite{svg2,SpargeAttention,Sparse-vDiT}.
Static methods avoid this cost by fixing sparse patterns offline, but rely on heuristics that fail to capture the content-adaptive structure of video attention \cite{svg,adaspa}.
Consequently, current methods typically sacrifice either inference efficiency or generation fidelity.

In this work, we argue that this dilemma stems from a flawed premise:
sparse attention patterns in Video DiTs are assumed to be purely input-dependent and transient.
Instead, we make the following empirical discovery:

In Video Diffusion Transformers, although individual attention activations vary across inputs, the union of high-mass attention regions for each head rapidly converges to a stable structure across prompts.
This empirical regularity indicates that sparsity is not transient, but reflects an underlying structural constraint of the model.

We therefore identify an \textit{Intrinsic Sparse Topology}—a static, weight-encoded attention mask that is prompt-agnostic and scale-invariant, and emerges directly from the pre-trained parameters.

This observation reframes sparse attention in Video DiTs: sparsification fundamentally decomposes into two orthogonal questions—\textit{where} attention is permitted to occur (topology), and \textit{how much} sparsity can be tolerated (sensitivity). Failing to disentangle these aspects, as in prior work, inevitably leads to either inefficiency or fidelity degradation.

Guided by this principle, we propose \textbf{ScalingAttention}, a training-free
acceleration framework that decouples sparse pattern discovery from adaptive
sparsity control.
ScalingAttention consists of three tightly co-designed components: 
\textbf{WEST} (Weight-Encoded Sparse Topology) extracts a robust,
resolution-scalable block-sparse prior mask via offline profiling; 
\textbf{FAST} (Fidelity-Aware Sensitivity Tuning) modulates head-wise sparsity in a fidelity-aware manner across the diffusion process;
and a hardware-aligned bit-wise block-sparse kernel (\textit{crm kernel}) translates structured sparsity into efficient block-level execution~\cite{flexattention,FA1}.

Extensive experiments on Wan2.1-1.3B and Wan2.1-14B demonstrate that ScalingAttention achieves
substantial end-to-end speedups while preserving generation quality,
establishing a new Pareto frontier between efficiency and fidelity.
In particular, our method delivers up to 1.90$\times$ end-to-end speedup while maintaining high visual fidelity.
Beyond acceleration, our findings reveal an intrinsic structural property of attention in Video Diffusion Transformers, offering a principled perspective on how sparsity emerges from pre-trained weights.

\begin{figure}[t]
  \centering
  \includegraphics[width=0.8\linewidth]{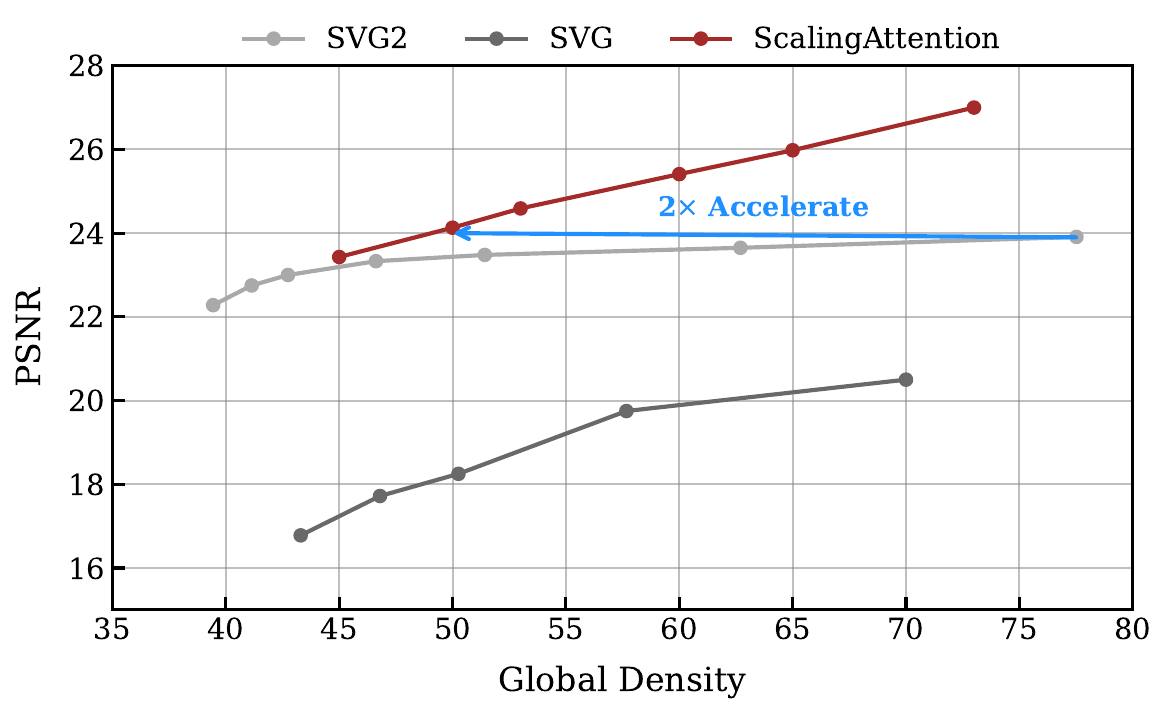}
  \vspace{-7pt} 
  \caption{\textbf{Performance Comparison.} 
We compare ScalingAttention with SVG and SVG2 on Video DiTs. 
At comparable PSNR, ScalingAttention uses up to $\mathbf{2\times}$ fewer attention FLOPs (i.e., lower \emph{density}) than SVG2, demonstrating a superior efficiency--fidelity trade-off.
\emph{Here, density is defined as $1-\text{Sparsity}$ and represents the fraction of active attention blocks; this convention is used consistently throughout the paper.}}
  \label{fig:performance_comparison}
  \vspace{-2.5em} 
\end{figure}

\section{Related Work}
\label{sec:relatedwork}

\subsection{Visual Generative Models}

Diffusion Transformers (DiTs) have evolved from factorized designs to full 3D architectures. Early video DiTs like Latte \cite{Latte} and OpenSora \cite{opensora} decoupled spatial and temporal attention to reduce computational cost. However, to model complex motion and maintain global consistency, state-of-the-art models (e.g., Wan2.1 \cite{wan2025wanopenadvancedlargescale}, HunyuanVideo \cite{kong2025hunyuanvideosystematicframeworklarge}) have shifted to full 3D attention, treating video as a unified token sequence \cite{sora, Qwen-image}. While this maximizes modeling capacity, it induces a quadratic complexity $O(N^2)$ that renders high-resolution generation computationally prohibitive, necessitating efficient attention mechanisms.

\subsection{Sparse Attention for Video DiTs}
Existing acceleration strategies largely fall into two categories. Dynamic Methods (e.g., SVG2 \cite{svg2}, AdaSpa \cite{adaspa}) prune tokens based on runtime activation magnitudes. While accurate, they require calculation-heavy selection steps and produce irregular memory patterns that hinder Tensor Core utilization on GPUs. Static Methods (e.g., Radial \cite{Radial}, PARO \cite{paro}) apply fixed sparsity masks (e.g., sliding windows or dilated patterns \cite{Longformer,Deepcache}) independent of input content. Although hardware-friendly, these heuristics lack the flexibility to capture object-specific interactions, often degrading visual quality. ScalingAttention bridges this gap by discovering static yet adaptive sparse patterns encoded in the model weights.

\section{Motivation}
\label{motivation}

\begin{figure*}[t]
  \centering
  \includegraphics[width=1\textwidth]{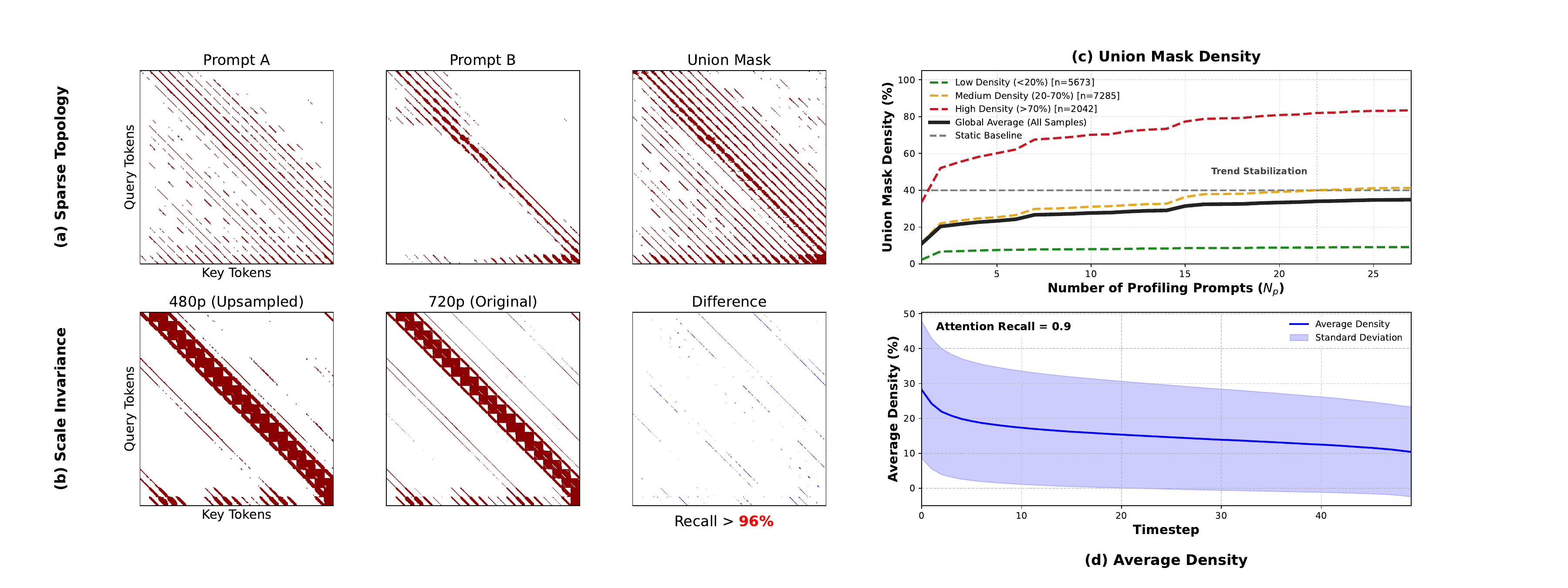}
  \vspace{-1.0em} 
  
  \caption{\textbf{Empirical Discovery of Intrinsic Sparse Topology.} 
  We reveal that attention sparsity in Video DiTs is governed by a static, weight-encoded structure rather than transient input variations.
  (a) \textbf{Sparse Topology (Top-Left):} While individual prompts (A vs. B) activate distinct sub-regions, their union converges to a stable, prompt-agnostic boundary, revealing a latent topology.
  (b) \textbf{Scale Invariance (Bottom-Left):} This topology is semantic: a mask extracted from a 480p proxy aligns with the native 720p map ($>96\%$ recall), enabling efficient low-res profiling.
  (c) \textbf{Calibration Stability (Right):} The union mask density saturates rapidly within $\sim$20 prompts, confirming that the intrinsic topology can be robustly extracted offline.}
  \label{fig:motivation_analysis}
  
  \vspace{-1.0em}
\end{figure*}

Despite the growing interest in sparse attention for accelerating Video Diffusion Transformers (DiTs), existing approaches remain constrained by a transient view of sparsity.
Prior methods~\cite{svg2} assume that attention patterns are purely input-dependent, leading to a persistent dichotomy: static methods impose rigid heuristics that fail to capture semantic structure, while dynamic methods incur prohibitive overhead from online identification of critical tokens.

In this work, we challenge this premise.
We reveal that while individual attention activations vary across inputs, the support regions of attention heads are largely predetermined by the model weights.
This reveals an \textit{Intrinsic Sparse Topology}—a static, weight-encoded structure that is stable across inputs and scale-invariant.
Based on this discovery, we identify three key observations that bridge the gap between algorithmic sparsity and hardware-efficient acceleration.

\paragraph{Observation 1: Sparse attention exhibits an intrinsic, weight-encoded topology.}
Contrary to the view that sparse patterns are purely input-dependent, we observe that each attention head possesses a latent sparse topology determined by its pre-trained weights. As visualized in Figure~\ref{fig:motivation_analysis}, this manifests in three key properties:
\begin{itemize}
    \setlength\itemsep{-0.5em} 
    \item \textbf{Intrinsic Structure:} While individual prompts trigger distinct activations, the \textit{union} of their high-mass regions rapidly converges to a stable, prompt-agnostic boundary. We term this the \textbf{Intrinsic Sparse Topology}. It represents a head-specific support envelope, which is a static boundary encoded in the weights and selectively activated by inputs.
    \item \textbf{Asymptotic Stability:} The density of this union mask does not grow unboundedly but saturates at a low level. This reveals a general sparse hull that captures all structurally significant interactions across the data distribution.
    \item \textbf{Scale Invariance:} This topology scales consistently across resolutions: a mask extracted from a 480p proxy aligns with the native 720p map with $>96\%$ recall. It also remains highly consistent under large aspect-ratio changes, even when the same prompt produces markedly different layouts. Together, these results suggest a reusable structural prior encoded by the model rather than an input-specific layout. Further analysis is provided in Section~\ref{sec:sesitivity} and Appendix Figure~\ref{fig:appendix_aspect_ratio}.
\end{itemize}

\paragraph{Theoretical Interpretation: The Functional Support Envelope.} 
We attribute this stability to an emergent \textit{functional specialization} of attention heads during training. To generalize across heterogeneous scenes and massive training scales, each head is driven to encode parameters sufficient to support a consistent functional bias (e.g., motion coherence or object permanence). This process naturally induces a bounded support envelope, corresponding to the maximal extent over which the head may be activated across inputs. At inference time, a specific prompt activates only a subset of this pre-encoded support. This explains why the topology is static while activations are dynamic.

\paragraph{Observation 2: Sensitivity to sparsity evolves via entropy reduction.}
While the \textit{topology} (where to attend) is static, the \textit{tolerance} for approximation (how much to attend) is not. The diffusion process describes a trajectory from high-entropy noise to structured content \cite{diffentrophy}. As generation progresses, attention distributions become increasingly concentrated, and the model's sensitivity to sparsity correspondingly diminishes.
This manifests across two dimensions: \textbf{temporal denoising steps} and \textbf{spatial transformer layers}. Consequently, enforcing a uniform sparsity ratio is suboptimal: it wastes computation in robust later stages while degrading fidelity in sensitive early phases.

\paragraph{Observation 3: The Sparsity-Efficiency Gap.}
Finally, theoretical sparsity rarely translates to wall-clock speedup due to the \textbf{Granularity Mismatch}.
Modern GPUs rely on Tensor Cores for peak throughput, which demand dense, aligned data tiles (e.g., $128 \times 128$). Existing dynamic methods generate fine-grained, unstructured masks that cause \textbf{memory fragmentation} \cite{fa2,fa3}. The overhead of metadata lookups and non-coalesced loads often negates the computational savings \cite{flexattention}.
Therefore, a practical sparse solution must not only preserve semantic structure (Obs. 1) but also be structurally aligned with hardware tiling units.

\paragraph{Implication: Tripartite Co-design.}
These three observations dictate the architectural requirements of ScalingAttention:
(1) \textbf{Topology (WEST):} Extract the weight-encoded structure offline as a static prior to bypass runtime search;
(2) \textbf{Sensitivity (FAST):} Modulate sparsity levels dynamically to match the evolving fidelity requirements;
(3) \textbf{System (\textit{crm kernel}):} Enforces block-wise alignment and bit-wise topology encoding, enabling efficient Tensor Core execution.
This co-design transforms sparse attention from a conceptual approximation into a principled, high-performance acceleration framework.

\section{Methodology}
\label{sec:methods}
In this section, we present ScalingAttention, a sparse attention framework for efficient video DiTs. Building on two insights from Section~\ref{motivation}: (i) sparse topology is encoded in the model weights, and (ii) sparsity sensitivity evolves throughout the diffusion trajectory. Consequently, ScalingAttention integrates three components: 1) Weight-Encoded Sparse Topology for constructing static, reusable masks offline; 2) Fidelity-Aware Sparsity Tuning for adaptive, fidelity-guided sparsity control throughout the generation process; and 3) an efficient bit-wise block-sparse attention kernel that translates theoretical FLOPs reduction into end-to-end inference acceleration with minimal runtime overhead.
\begin{figure*}[t]
  \centering
  \includegraphics[width=\textwidth]{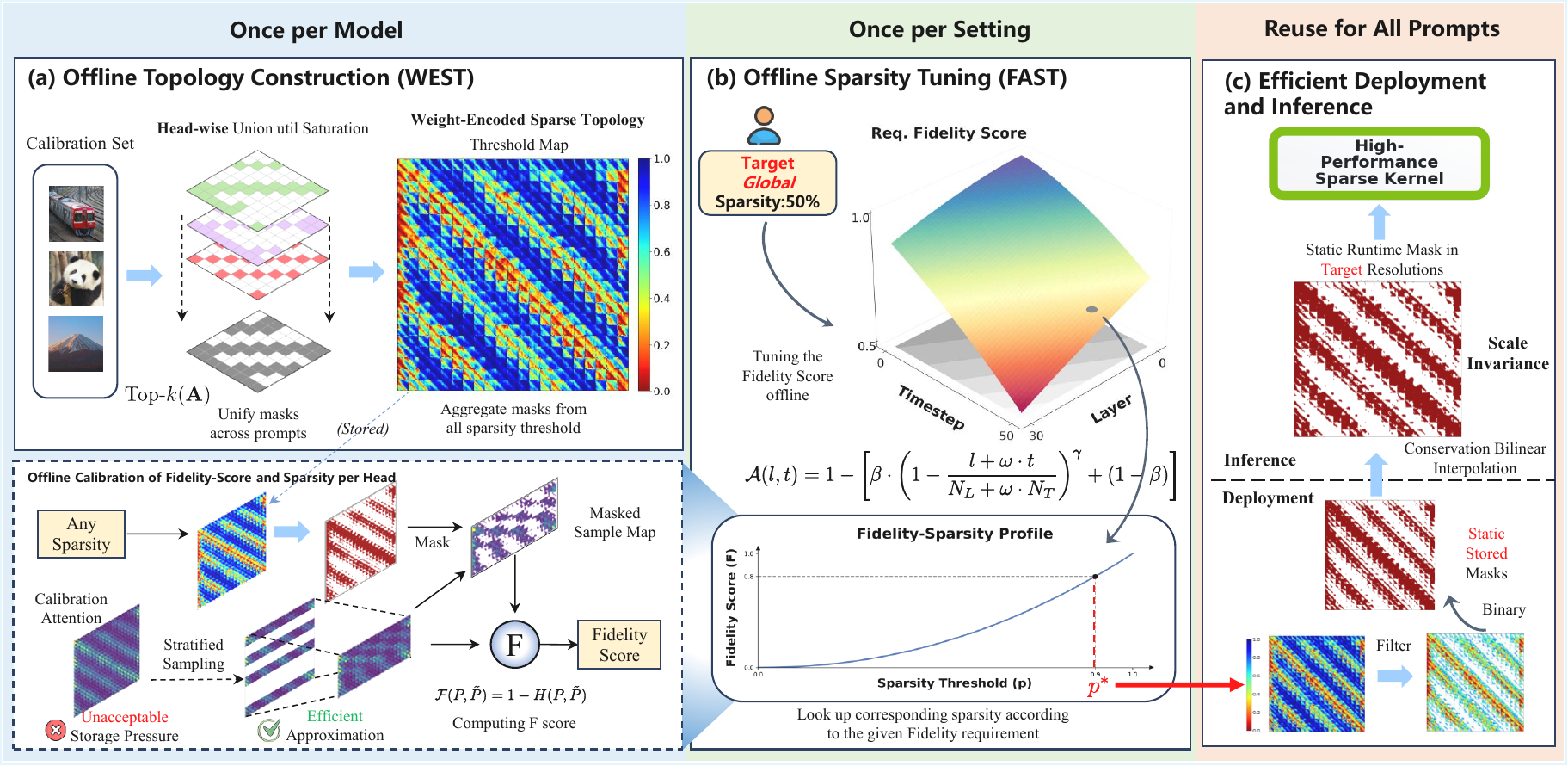}
  \vspace{-1.5em}
  \caption{\textbf{Overview of the ScalingAttention Framework.}
  Our method decouples sparse attention into three phases:
  \textbf{(a) Once per Model (WEST):} We aggregate intrinsic attention patterns from a calibration set to construct a static \textit{Threshold Map}, which encodes the complete sparsity hierarchy offline.
  \textbf{(b) Once per Setting (FAST):} Given a global sparsity target, we modulate a spatio-temporal fidelity surface to determine head-specific thresholds. Using \textit{Stratified Sampling} and a pre-computed \textit{Fidelity-Sparsity Profile}, we map these requirements to optimal thresholds $p^*$ via efficient lookup.
  \textbf{(c) Reuse for All Prompts:} The resulting static masks are resolution-scalable via \textit{Conservative Bilinear Interpolation} and executed by proposed \textit{crm kernel}.}
  \label{fig:method}
  \vspace{-1.0em}
\end{figure*}

\subsection{Weight-Encoded Sparse Topology (WEST)}
\label{sec:west}

WEST extracts a static sparse topology from pre-trained weights. Instead of storing discrete masks, we construct an offline \emph{Threshold Map} that compactly indexes block stability, enabling $\mathcal{O}(1)$ mask generation at inference.

\subsubsection{Block-wise Topology Extraction}
\label{sec:block_extraction}

Given a calibration set, we compute the attention matrix $\mathbf{A}_i$ for each head. To align with hardware tiling (Sec.~\ref{sec:crm_attention}), we partition $\mathbf{A}_i$ into $B \times B$ blocks ($B=128$) and aggregate scores into a block significance matrix $\mathbf{S}_i$:
\begin{equation}
    \mathbf{S}_i^{(u,v)} = \frac{1}{B^2} \sum_{p \in \mathcal{B}_u} \sum_{q \in \mathcal{B}_v} \mathbf{A}_i^{(p,q)}.
    \label{eq:block_pool}
\end{equation}
where $\mathcal{B}_u$ and $\mathcal{B}_v$ denote the token indices of the $u$-th query block and
$v$-th key block, respectively.
This aggregation filters token-level noise while preserving stable, hardware-friendly structural dependencies.

\subsubsection{Threshold Map Construction}
\label{sec:topology_union}

To capture the \textit{intrinsic topology} generalizing across inputs, we define the \textit{Threshold Map} $\mathbf{T}$. For each block $(u,v)$, $\mathbf{T}^{(u,v)}$ records the minimum sparsity threshold $p$ required for inclusion in the union of active regions across all calibration prompts:
\begin{equation}
    \scalebox{0.92}{$\mathbf{T}^{(u,v)} = \min \{ p \mid \exists i \in [1, M], (u,v) \in \text{Top-}k(\mathbf{S}_i, p) \}. $}
\end{equation}
This representation encodes the full sparsity hierarchy. At inference, a binary mask for any target density $p$ is instantiated via a lightweight comparison $\mathbf{M}_p = \mathbb{I}(\mathbf{T} \le p)$, effectively decoupling topology discovery from dynamic sparsity control.
We visualize the resulting \textit{threshold map} and the mask construction process in Figure~\ref{fig:method}. Additional visualizations across layers/heads are provided in Appendix~\ref{app:threshold_maps}.

\subsection{Fidelity-Aware Sparsity Tuning (FAST)}
\label{sec:fast}

FAST determines the optimal sparsity $p(l,t)$ for each layer $l$ and timestep $t$ by regulating the divergence between dense and sparse attention. Unlike static strategies, FAST adheres to a fidelity-first principle: it dynamically allocates computation budget to early denoising steps and shallower layers where error accumulation makes the model most sensitive (Observation 2), ensuring that aggressive sparsification in later stages does not compromise generation quality.

\subsubsection{Hellinger Distance as a Fidelity Metric}
\label{sec:hellinger_distance}

\begin{figure}[t]
    \centering
    \includegraphics[width=0.8\linewidth]{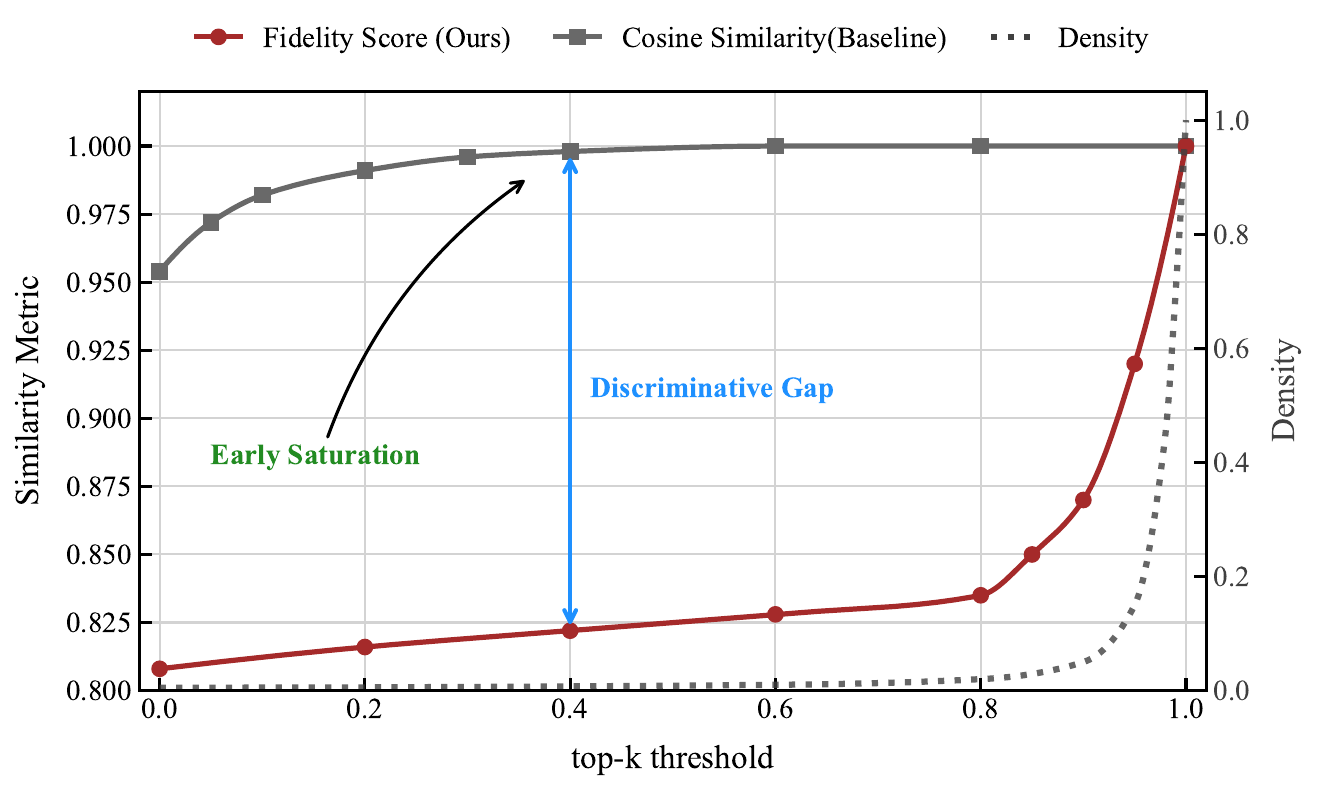}
    \vspace{-0.5em}
    \caption{
    \textbf{Metric Comparison.}
    Cosine similarity (orange) saturates prematurely as sparsity increases, failing to capture structural loss. The proposed Fidelity Score (blue) maintains a smooth, monotonic response, enabling precise calibration.
    }
    \label{fig:similarity_metrics}
    \vspace{-1em}
\end{figure}

Adaptive sparsity control requires a reliable metric to quantify the discrepancy
between dense attention distributions $\mathbf{P}$ and their sparse approximations $\tilde{\mathbf{P}}$.
Conventional measures such as cosine similarity are ill-suited for attention maps,
which are typically high-dimensional and heavy-tailed.
As shown in Figure~\ref{fig:similarity_metrics}, cosine similarity
saturates prematurely, approaching $1.0$ even when structurally critical
tokens have been pruned.
This occurs because cosine similarity measures only angular alignment:
once attention mass collapses onto a few dominant entries, sparse and
dense attention maps remain nearly collinear, rendering cosine similarity
insensitive to structural distortion \cite{cosim}.

We therefore adopt a fidelity metric based on the \textit{Hellinger distance} \cite{divergence}.
For two discrete distributions $P$ and $Q$, it is defined as:
\begin{equation}
    H(P, Q) =
    \frac{1}{\sqrt{2}}
    \sqrt{\sum_{i=1}^{n} \left(\sqrt{p_i} - \sqrt{q_i}\right)^2}.
\end{equation}
The square-root transformation attenuates extreme values,
making the metric more sensitive to changes in distributional mass.

And compared to KL divergence, Hellinger distance is symmetric, bounded, and well-defined under zero-mass entries, which are ubiquitous in sparse attention \cite{KL}. Compared to total variation, its square-root geometry provides smoother sensitivity under progressive sparsification.

For dense attention $P$ and its sparse approximation $\tilde{P}$,
we define the \textbf{Fidelity Score} as:
\begin{equation}
    \mathcal{F}(P, \tilde{P}) = 1 - H(P, \tilde{P}).
\end{equation}
As illustrated in Figure~\ref{fig:similarity_metrics}, the Fidelity Score
maintains a smooth, monotonic response across the entire sparsity range,
enabling reliable threshold calibration for FAST.

\subsubsection{Spatial-Temporal Fidelity Allocation}
\label{sec:fidelity_allocation}

To map the model’s evolving tolerance to specific sparsity levels,
we define a parametric target fidelity function
$\mathcal{A}(l,t)$ over transformer layer $l$ and denoising timestep $t$:
\begin{equation}
\scalebox{0.99}{$
\mathcal{A}(l,t)=1-\left[\beta\left(1-\frac{l+\omega t}{N_L+\omega N_T}\right)^{\gamma}+(1-\beta)\right]
$}
\label{eq:spatio_temporal}
\end{equation}
where $N_L$ and $N_T$ denote the total number of layers and timesteps.
The parameter $\omega$ balances spatial and temporal contributions,
while $\beta$ (baseline fidelity) and $\gamma$ (transition curvature)
are optimized offline to satisfy the given constraint
(e.g., $50\%$ overall global sparsity), rather than being manually tuned.

During offline profiling, we characterize the fidelity-sparsity relationship for each attention head
under its corresponding WEST topology.
Prior to inference, given a global density target,
FAST determines a consistent set of head-specific sparsity thresholds
that collectively satisfy the global constraint.
For each head at position $(l,t)$,
we select the maximum sparsity threshold
that satisfies:
\begin{equation}
    \mathcal{F}(P, \tilde{P}) \ge \mathcal{A}(l,t).
\end{equation}
Combined with the \textit{threshold map} from Sec.~\ref{sec:west},
this yields static sparse masks for inference without runtime adaptation.
The overall workflow is illustrated in Figure~\ref{fig:method}, and the complete mask generation procedure is detailed in
Algorithm~\ref{alg:mask_gen} in the Appendix.

\subsubsection{Efficient Profiling via Stratified Sampling}
\label{sec:stratified_sample}

Directly computing fidelity on full calibration $N \times N$ attention maps offline
would incur prohibitive storage and computational overhead during profiling.
To make threshold calibration practical,
we employ a \textit{block-wise stratified sampling strategy} to the full calibration attention map.

Instead of evaluating all rows,
we partition the attention map into uniform row blocks
and randomly sample a small number of rows from each block.
This ensures coverage of both high- and low-importance regions. Figure~\ref{fig:method} visualizes the \textit{stratified sampling} procedure.
Empirically, sampling only $16$ rows
(approximately $0.05\%$ of the full map)
suffices to approximate the true Hellinger distance with negligible error
(see sensitivity analysis in Sec.~\ref{sec:sesitivity_sample}).
This reduces profiling cost by orders of magnitude
while preserving calibration accuracy.

\subsection{Efficient System--Algorithm Co-design}
\label{sec:system_codesign}

While Sections \ref{sec:west} and \ref{sec:fast} establish the theoretical efficiency of ScalingAttention, 
to translate these algorithmic gains into practical speedups,
we further address two system-level challenges that arise in deployment:
(i) supporting variable resolutions without quadratic storage overhead and re-profiling,
and (ii) executing diverse sparse patterns efficiently on modern GPUs.
We resolve both through a cohesive system--algorithm co-design,
which enables scalable deployment and hardware-efficient execution.

\subsubsection{Resolution Scalability}
\label{sec:resolution_scalability}

Since sparse patterns are determined by model weights rather than sequence length,
ScalingAttention constructs sparse topology only once at a base resolution
(e.g., $480 \times 720$) and reuses it across scales.
At inference, the base \textit{threshold map} is resized via \textit{conservative bilinear interpolation}
with a ceiling rule:
\begin{equation}
    \mathbf{R}_{\text{target}}^{(u,v)} =
    \left\lceil
    \Phi\!\left(
        \mathbf{R}_{\text{base}},
        \frac{u}{s_h}, \frac{v}{s_w}
    \right)
    \right\rceil ,
\end{equation}
ensuring that any overlap with active regions is preserved.
This enables arbitrary-resolution inference with constant storage cost.

\subsubsection{CRM Kernel: Bit-wise Block-Sparse Attention Execution}
\label{sec:crm_attention}

Efficient block-sparse attention requires a runtime representation that supports both compact storage and fast iteration over active KV blocks \cite{deepspeed}.
Index-based formats (e.g., CSR) incur non-negligible metadata traffic and irregular memory access,
while dense boolean masks suffer from excessive space overhead \cite{sparsegpu, FA1,fa2}.

The \textit{crm kernel} adopts a \emph{Compressed Row Mask (CRM)} representation to encode sparse topology,
representing each attention row as a compact bitmask stored in aligned \texttt{uint32} arrays,
where each bit indicates the activation of a KV block.

This design enables a single \texttt{uint32} load to decode the status of 32 blocks simultaneously.
With a block size of $128 \times 128$, even long sequences (e.g., 256k tokens) require only tens of mask loads per row (64 loads for 256k tokens), making mask overhead negligible relative to attention computation.

Active KV blocks are traversed using native GPU bitwise operations
(e.g., bit scan and bit clear),
allowing constant-time advancement without explicit index decoding.
This iterator seamlessly replaces the dense block iterator in FlashAttention,
yielding a block-sparse variant that preserves dense Tensor Core computation \cite{fa3}.

Pseudocode and detailed kernel-level evaluations are provided in the Appendix~\ref{app:crm_code}.

\section{Experiments}
\label{sec:5}

\subsection{Experimental Setup}

\paragraph{Models and Tasks.}
We evaluate ScalingAttention on state-of-the-art open-source Video Diffusion Transformers,
including Wan2.1-T2V-1.3B, Wan2.1-I2V-14B \cite{wan2025wanopenadvancedlargescale}, and HunyuanVideo \cite{kong2025hunyuanvideosystematicframeworklarge}.
Experiments are conducted at $480$p for Wan2.1-T2V-1.3B and HunyuanVideo, and at $720$p for Wan2.1-I2V-14B.
For cross-model validation on HunyuanVideo, we evaluate the full WEST+FAST pipeline at 55\% global density on a PenguinVideo Benchmark subset that covers diverse semantic aspects.
After 3D-VAE tokenization, Wan2.1-1.3B processes 21 frames with 1,560 tokens, Wan2.1-14B processes 3,600 tokens while HunyuanVideo processes 33 frames with 1350 tokens per frame.

\paragraph{Metrics.}
We assess reconstruction fidelity using PSNR, SSIM, and LPIPS, and evaluate perceptual video quality with VBench \cite{vbench,vbench2}.
Efficiency is measured by \textit{global density}, defined as the ratio of sparse attention FLOPs to full attention FLOPs across all layers and timesteps.

\paragraph{Baselines.}
We compare against Sparse VideoGen (SVG) \cite{svg} and Sparse VideoGen2 (SVG2) \cite{svg2}, using their official configurations and identical prompts for fair comparison.

\paragraph{Implementation Details.}
ScalingAttention is implemented with custom CUDA kernels and benchmarked on high-end GPU with large memory capacity, with GPU clock locked to 1545\,MHz to avoid potential throttling.
For WEST, sparse topologies are constructed using $N_p=27$ diverse calibration prompts.
For FAST, fidelity profiling uses a single prompt, which we find sufficient in practice.
Unless otherwise specified, our method applies sparsity across all layers and denoising steps \textbf{without any dense warm-up}, whereas prior methods rely on a warm-up phase.
Additional details are provided in the Appendix.

\subsection{Quality Evaluation}

\begin{table*}[t]
    \centering
    \renewcommand{\arraystretch}{1.2} 
    \setlength{\tabcolsep}{5.5pt}
    
    \caption{Quality and efficiency benchmarking results of ScalingAttention and baselines on three representative Video DiTs (Wan~2.1~1.3B, Wan~2.1~14B, and HunyuanVideo). ScalingAttention consistently achieves superior quality--speedup trade-offs.}
    \label{tab:results}
    
    \small 
    \begin{tabular}{ll|ccccc|cc}
        \toprule[1.2pt]
        \textbf{Model} & \textbf{Config} & \textbf{PSNR} $\uparrow$ & \textbf{SSIM} $\uparrow$ & \textbf{LPIPS} $\downarrow$ & \textbf{Smoothness} $\uparrow$ & \textbf{Consistency} $\uparrow$ & \textbf{Density} $\downarrow$ & \textbf{Speedup} $\uparrow$ \\
        \midrule[1pt]
        
        \textbf{Wan 2.1} & \textit{1.3B, 480P, T2V} & - & - & - & 98.46 & 96.07 & 100\% & 1$\times$ \\
        \hline
        & SVG & 19.71 & 0.7314 & 0.2684 & 97.87 & 92.58 & 50\% & 1.27$\times$ \\
        & SVG2 & 24.44 & 0.8387 & 0.1249 & 98.25 & 92.46 & 50\% & 1.31$\times$ \\
        \rowcolor{rowblue} 
        & \textbf{Ours} & \textbf{26.61} & \textbf{0.8734} & \textbf{0.0986} & 98.31 & 93.16 & 50\% & \textbf{1.65}$\times$ \\
        \hline
        
        \textbf{Wan 2.1} & \textit{14B, 720P, I2V} & - & - & - & 98.96 & 98.46 & 100\% & 1$\times$ \\
        \hline
        & SVG & 22.34 & 0.7385 & 0.1597 & 98.62 & 95.54 & 52\% & 1.47$\times$ \\
        & SVG2 & 23.33 & 0.7474 & 0.1431 & 98.50 & 95.75 & 52.5\% & 1.56$\times$ \\
        \rowcolor{rowblue} 
        & \textbf{Ours} & \textbf{24.15} & \textbf{0.7778} & \textbf{0.1191} & 98.65 & 96.00 & 52.5\% & \textbf{1.90}$\times$ \\
        \hline

        \textbf{HunyuanVideo} & \textit{480P, T2V} & - & - & - & 98.87 & 91.59 & 100\% & 1$\times$ \\
        \hline
        & SVG & 26.4696 & 0.8592 & 0.1125 & 98.94 & 91.76 & 55\% & 1.68$\times$ \\
        & SVG2 & 28.3100 & 0.8853 & 0.0860 & 98.89 & 91.26 & 55\% & 1.69$\times$ \\
        \rowcolor{rowblue}
        & \textbf{Ours} & \textbf{28.8844} & \textbf{0.8996} & \textbf{0.0756} & 98.85 & 91.45 & 55\% & \textbf{1.73}$\times$ \\
        
        \bottomrule[1.2pt]
    \end{tabular}
    \vspace{-0.1in} 
\end{table*}

We qualitatively validate our method by visualizing attention
maps. As shown in Figure~\ref{fig:motivation_analysis}, we analyze maps from various attention heads in Wan2.1 using prompts from VBench.
Despite the diversity of sparse patterns, they exhibit remarkable topological consistency across prompts (i.e., different \textit{threshold maps} in Appendix~\ref{app:threshold_maps}). Additionally, benefiting from resolution
scalability, the same sparse mask can be reused across different resolutions. This allows for flexible resolution and
context changes with lightweight pre-processing at a lower resolution.

We present a quantitative comparison against state-of-the-art
baselines in Table~\ref{tab:results}. ScalingAttention consistently establishes a new Pareto frontier, outperforming all baseline methods in reconstruction fidelity (PSNR, SSIM, LPIPS) while achieving the highest speedup due to
its co-design. Specifically, ScalingAttention achieves a PSNR of 26.61 dB on Wan2.1-1.3B, 24.15 dB on Wan2.1-14B, and 28.88 dB on HunyuanVideo at substantially reduced density, demonstrating its effectiveness in preserving fine details and temporal quality while removing redundant computation. On HunyuanVideo, the results are measured on a PenguinVideo Benchmark subset, where our method also remains competitive on the two retained perceptual metrics, Smoothness and Consistency, under the same evaluation protocol.

Detailed visual comparisons on HunyuanVideo are deferred to Appendix Figure~\ref{fig:hunyuan_visual_comparison}. The sparse outputs preserve dense-level appearance while maintaining a sparsity of nearly 50\% in Table~\ref{tab:results}, demonstrating nearly lossless video quality and acceleration benefits.

\subsection{Efficiency Evaluation}

\begin{figure*}[t]
    \centering
    \includegraphics[width=\linewidth]{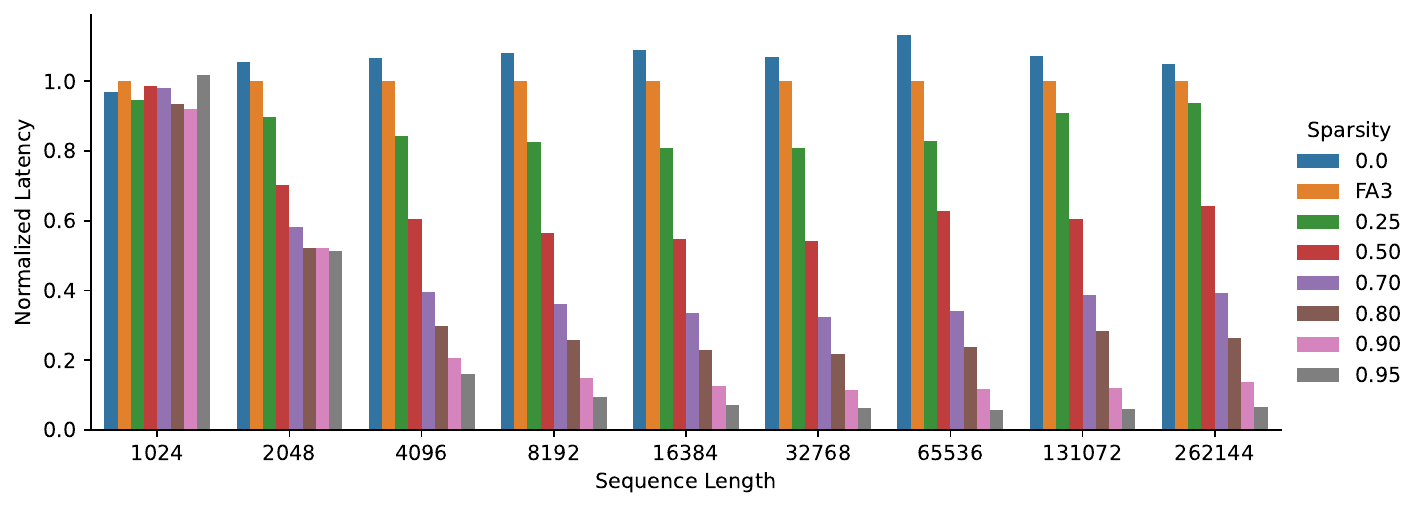}
    \vspace{-1.5em} 
    \caption{\textbf{Kernel-Level Efficiency Benchmarking.} We compare the normalized latency of our \texttt{\textit{crm kernel}} kernel against FlashAttention-3 (FA3) across sequence lengths from 1K to 262K. The orange bar represents the FA3 baseline (1.0). Even at 0\% sparsity (Blue), our kernel incurs minimal overhead ($<10\%$) due to efficient bit-mask loading. As sparsity increases (Green to Grey), latency drops significantly, demonstrating linear scalability.}
    \label{fig:speedup_sparsity}
    \vspace{-1.0em} 
\end{figure*}

\paragraph{Efficient Block-Sparse Kernel.}
To demonstrate the efficiency of the \textit{crm kernel}, we benchmark its latency against FlashAttention-3~\cite{fa3} across varying sparsity levels and sequence lengths using a fixed block size of 128, as shown in Figure~\ref{fig:speedup_sparsity}.
We analyze performance from two aspects:

(1) \textbf{Minimal Overhead:} We evaluate implementation overhead in the fully dense setting (0
As shown in Figure~\ref{fig:speedup_sparsity} (with short-sequence details in Appendix~\ref{app:fig_short_overhead}), the \textit{crm kernel} incurs less than 10
These results confirm that the \textit{crm kernel} is an efficient block-sparse extension of FlashAttention, supporting high-resolution video generation with minimal cost.

(2) \textbf{Scalable Speedup:} As sparsity increases, the \textit{crm kernel} effectively converts theoretical FLOPs reduction into wall-clock speedup: at 50\% sparsity, latency is reduced by 40--60\%, and at extreme sparsity ($>90\%$), it achieves over $10\times$ acceleration, demonstrating strong scalability for long-context modeling.

\textbf{End-to-End Speedup Evaluation.}
To demonstrate the end-to-end efficiency of ScalingAttention, we report several metrics, including global density and speedup. In Table~\ref{tab:results}, ScalingAttention achieves an average speedup of 1.76$\times$, 
while maintaining superior generation quality.
Specifically, compared to SVG2, ScalingAttention achieves both higher fidelity (PSNR of 26.61 dB) and greater acceleration (1.65$\times$) on Wan2.1-1.3B at 50
Moreover, on Wan2.1-14B, ScalingAttention delivers a \textbf{1.90}$\times$ speedup at 52.5\% density while still attaining the highest PSNR among all methods. The same trend holds on HunyuanVideo, where ScalingAttention reaches 1.73$\times$ speedup at 55\% density while also achieving the best reconstruction metrics in Table~\ref{tab:results}.

\subsection{Sensitivity Analysis}
\label{sec:sesitivity}
We evaluate the robustness of ScalingAttention against hyperparameter variations to validate its stability on the efficiency-accuracy Pareto frontier.

\textbf{Impact of Calibration Set Size ($N_p$).} 
\label{sec:sesitivity_prompts}
We assess the sensitivity of WEST to the number of profiling prompts. We construct the calibration set $\mathcal{D}_{cal}$ by sampling representative prompts across diverse semantic dimensions (e.g., camera motion, object interactions) to ensure sufficient coverage. 
We observe that the topological structure saturates rapidly. As shown in Figure~\ref{fig:motivation_analysis}(c), the union mask density stabilizes even when aggregating $N_p=27$ prompts, indicating that individual attention patterns are consistently confined within a compact static topology. 
Empirically, a lightweight calibration set of as few as 8--10 diverse prompts is sufficient to approximate the optimal topology with negligible performance loss.

\textbf{FAST Robustness to the Profiling Prompt.}
FAST does not discover semantic topology; that role is already handled by WEST using diverse calibration prompts. Instead, FAST only profiles how aggressively each head can be sparsified under a fixed WEST-derived \textit{Threshold Maps} prior. 

To test whether a single profiling prompt is sufficient, we vary the prompt used for threshold estimation while keeping the same underlying sparse topology and fixed target density on Wan2.1-1.3B. Specifically, we compare four semantically distinct prompt groups (\textit{action}, \textit{scenery}, \textit{object}, and \textit{sci-fi}) against an aggregate reference formed by averaging their selected scale indices. Threshold-level statistics initially suggest noticeable variation: the mean per-head activated index range is 6.90, and the maximum range reaches 48. However, these threshold shifts do not translate into substantial mask differences, because the attention density curves are typically flat in the long-tail region and nearby thresholds often induce nearly identical token masks. As shown in Table~\ref{tab:prompt_sensitivity}, every single prompt remains close to the aggregate reference, with IoU above 91\% and density deviation within 1.89\%. The average density gap is below 1.3\%, and the \textit{sci-fi} prompt provides the closest approximation, with only 0.24\% density gap and 94.53\% IoU. This result supports the practical use of a single profiling prompt for FAST, with negligible impact on the recovered sparse support and generated videos.

\begin{table}[t]
    \centering
    \renewcommand{\arraystretch}{1.1}
    \setlength{\tabcolsep}{4pt}
    \caption{\textbf{FAST robustness to the profiling prompt.} With fixed WEST topology, different profiling prompts lead to only minor density variation and high mask overlap with the aggregate reference.}
    \label{tab:prompt_sensitivity}
    \small
    \begin{tabular}{lcccc}
        \toprule
        \textbf{Profiling Prompt} & \textbf{Density} & $\Delta$ \textbf{Density} & \textbf{IoU} & \textbf{Scale Err.} \\
        \midrule
        Aggregate ref. & 39.05\% & -- & -- & -- \\
        sci-fi & 38.80\% & 0.24\% & 94.53\% & 1.62 \\
        action & 38.01\% & 1.03\% & 91.10\% & 2.81 \\
        scenery & 37.22\% & 1.83\% & 91.90\% & 2.80 \\
        object & 40.94\% & 1.89\% & 93.63\% & 1.98 \\
        \bottomrule
    \end{tabular}
\end{table}

\textbf{Sensitivity to Profiling Sampling Ratio.} 
\label{sec:sesitivity_sample}
We investigate the trade-off between estimation precision and profiling overhead by varying the row sampling ratio. 
Empirical results indicate that \textit{block-wise stratified sampling} with only 16 rows ($\sim$0.05\% of the sequence) is sufficient.
Detailed results are provided in \textbf{Appendix~\ref{app:sample_ratio}}.
This minimal subset yields a correlation of $>0.999$ with the ground truth derived from the full attention map. 
Consequently, we adopt $N=16$ to minimize offline profiling costs without compromising calibration robustness.

\textbf{Offline Cost and Storage.}
We further quantify the one-time setup cost of WEST and FAST on HunyuanVideo-13B using a GPU. WEST uses about 10 dense calibration prompts, whereas FAST requires only one dense generation pass for Hellinger-distance-based fidelity profiling. As summarized in Table~\ref{tab:offline_cost}, the dominant cost lies in WEST prompt sampling, while threshold-map construction, density search, and target-mask generation are substantially smaller. Although the intermediate tensors are large, these costs are amortized across later inference runs because threshold maps and target-density masks are reusable. For reference, generating a single 5\,s HunyuanVideo-720P video already takes nearly 2300\,s on one GPU, making the one-time profiling pipeline practical when reused. Moreover, the runtime overhead of the CRM kernel remains below 10\% in the dense setting and only 4.8\% on long sequences, as shown in Figure~\ref{fig:speedup_sparsity}.

\begin{table}[t]
    \centering
    \renewcommand{\arraystretch}{1.12}
    \setlength{\tabcolsep}{4pt}
    \caption{\textbf{One-time offline cost of WEST and FAST on HunyuanVideo-13B.} Threshold maps and target-density masks are reusable across subsequent inference runs.}
    \label{tab:offline_cost}
    \small
    \begin{tabular}{p{0.49\columnwidth}cc}
        \toprule
        \textbf{Stage} & \textbf{Time (s)} & \textbf{Storage} \\
        \midrule
        WEST sampling (one prompt) & 2228 & 17 GB \\
        FAST sampling (one prompt) & 340 & 96 GB \\
        Threshold map + Fidelity Score construction & 759 & 8.6 GB \\
        Density search & 330 & -- \\
        One target-density mask & 118 & 8.2 GB \\
        \bottomrule
    \end{tabular}
\end{table}

\textbf{Robustness Under Aspect-Ratio Shift.}
\label{sec:aspect_ratio_shift}
To test whether the discovered topology remains stable under large geometric changes, we additionally evaluate Wan2.1 using $480 \times 832$ and $832 \times 480$ inputs. Even for the same prompt, these settings can produce substantially different semantic layouts. Nevertheless, the resulting attention distributions still share nearly the same sparse support pattern and maintain high recall. This cross-aspect-ratio consistency further supports that WEST captures a reusable support envelope determined primarily by the model rather than by a specific input layout. Detailed visualizations of the masks across different aspect ratios are deferred to Appendix Figure~\ref{fig:appendix_aspect_ratio}. In practice, we adopt a mixed calibration strategy that samples prompts from multiple aspect ratios to slightly improve coverage without introducing any runtime overhead.

\paragraph{Pareto Frontier Analysis.}
\label{sec:sesitivity_psnr}
To validate overall effectiveness, we benchmark ScalingAttention against baselines across a wide spectrum of computation budgets on Wan2.1-T2V-1.3B. 
As shown in Figure~\ref{fig:performance_comparison}, ScalingAttention consistently dominates the efficiency-quality trade-off, achieving superior fidelity at equivalent densities with higher speedups enabled by its zero-overhead design.
We identify two distinct regimes in the density-quality curve:

\begin{itemize}
    \item \textbf{Low-Density Regime ($<40\%$):} We observe a performance drop due to the \textit{granularity bottleneck}. The fixed block size of $128 \times 128$ limits the retention of fine-grained critical tokens under extreme sparsity. However, our static topology is orthogonal to token permutation methods (e.g., PARO, SVG2) \cite{paro,svg2}. As further validated by the compositional results in Table~\ref{tab:orthogonal_results}, combining ScalingAttention with token reordering can partially mitigate this bottleneck by clustering dispersed critical tokens into more hardware-friendly blocks.
    \item \textbf{High-Density Regime ($>40\%$):} Quality scales linearly with density. ScalingAttention significantly outperforms baselines that suffer from rigid heuristics or runtime overhead. Notably, our method achieves comparable PSNR to dense attention with only $\sim$50\% computation, validating the efficacy of FAST in preserving essential visual information.
\end{itemize}

\subsection{Compatibility with Orthogonal Techniques} 
\label{sec:orthogonal} 
We further evaluate whether ScalingAttention remains effective when composed with orthogonal acceleration techniques. Specifically, we apply our method to FastWan (3-steps)~\cite{vsa}, a highly sparse and distilled model, and additionally combine it with PARO-based token permutation~\cite{paro}. 

As shown in Table~\ref{tab:orthogonal_results}, ScalingAttention maintains stable performance in the distilled setting and benefits further from PARO, which improves PSNR from 24.61 to 24.92 at a nearly identical density. These results support two conclusions: first, ScalingAttention integrates seamlessly with external acceleration strategies without causing disproportionate error accumulation; second, token permutation provides a practical way to alleviate the coarse $128 \times 128$ granularity bottleneck by reorganizing sparse support into more favorable block layouts. 

\begin{table}[t] 
    \centering 
    \renewcommand{\arraystretch}{1.15} 
    \setlength{\tabcolsep}{3pt} 
    \caption{\textbf{Compositional evaluation with orthogonal techniques.} ScalingAttention remains effective in a 3-step distilled setting and benefits further from PARO.} 
    \label{tab:orthogonal_results} 
    \small 
    \begin{tabular}{lcccc} 
        \toprule 
        \textbf{Method} & \textbf{Density} & \textbf{PSNR} $\uparrow$ & \textbf{SSIM} $\uparrow$ & \textbf{LPIPS} $\downarrow$ \\ 
        \midrule 
        Ours + FastWan & 54.0\% & 24.61 & 0.8848 & 0.0732 \\ 
        \quad + PARO & 53.7\% & \textbf{24.92} & \textbf{0.8904} & \textbf{0.0680} \\ 
        \bottomrule 
    \end{tabular} 
\end{table}

\subsection{Ablation Study}
\label{sec:ablation}

\textbf{Effectiveness of FAST.} 
To isolate the contribution of our dynamic sparsity control, we compare ScalingAttention against a \textit{static uniform} baseline on Wan2.1-T2V-1.3B. Both methods use the same weight-encoded topology (WEST).
Crucially, to strengthen the baseline, we grant it a \textbf{30\% dense warm-up} phase. In contrast, ScalingAttention operates with \textbf{0\% warm-up}, applying adaptive sparsity throughout the entire generation process.

As detailed in Table~\ref{tab:ablation_sparsity} (\textbf{in Appendix~\ref{app:ablation_table}}), ScalingAttention consistently outperforms the baseline at all density levels. 
Notably, at a low density of 45\%, the static baseline performance collapses (17.83 dB), whereas our method maintains high fidelity (23.43 dB). 
This result confirms that static thresholds fail to capture the evolving sensitivity of the diffusion process, whereas FAST effectively aligns the computational budget with the model's intrinsic temporal dynamics.

\section{Conclusion}
We presented ScalingAttention, a system-algorithm co-designed framework that accelerates Video Diffusion Transformers by decoupling sparse topology discovery (WEST) from dynamic sensitivity modulation (FAST).
Our study reveals a strong inductive bias in Video DiTs: while attention activations are input-dependent, the sparse topology of each head is largely encoded in the model weights.

By exploiting this property with a hardware-aligned bitwise block-sparse kernel, ScalingAttention bridges the gap between theoretical sparsity and practical wall-clock acceleration, achieving state-of-the-art speedups without compromising generation quality.

We believe this perspective opens new opportunities for combining static structural priors with trainable modulation and token-level permutation mechanisms, further advancing efficient large-scale video generation. The compositional results in Section~\ref{sec:orthogonal} provide a concrete example of this direction.

\noindent\textbf{Limitations.} ScalingAttention has two primary limitations. 
First, the granularity bottleneck: to maximize Tensor Core utilization, our \textit{crm kernel} uses $128{\times}128$ blocks, which restricts fine-grained pruning and leads to diminished returns at extreme sparsity ($<40\%$). 
Second, static envelope redundancy: WEST extracts the union of active regions, inherently retaining more tokens than dynamic oracles for localized motion, a trade-off that future token permutation mechanisms could address.


\bibliography{example_paper}
\bibliographystyle{abbrvnat}

\newpage
\appendix
\onecolumn

\section{Algorithm for ScalingAttention Mask Generation}

We provide the complete algorithmic pipeline for sparse mask generation in ScalingAttention. The procedure consists of two offline phases:
(1) \textbf{Weight-Encoded Sparse Topology (WEST)}, which constructs a static threshold map and a fidelity profile from calibration data;
(2) \textbf{Fidelity-Aware Sparsity Tuning (FAST)}, which instantiates block-sparse attention masks for a given global density target.

\begin{algorithm}[h]
\small
    \caption{ScalingAttention: From Sampling to Static Mask Generation}
    \label{alg:mask_gen}
\begin{algorithmic}[1]
    \State {\bfseries Input:} Calibration dataset $\mathcal{D}$, Set of candidate thresholds $\mathcal{P} = \{p_1, \dots, p_K\}$
    \State {\bfseries Input:} Global target density $\rho_{target}$, Fidelity hyperparameters $(\alpha, \beta)$
    \State {\bfseries Output:} Set of Static Block Masks $\mathcal{M} = \{\mathbf{M}_{l,t,h}\}$ for all layers $l$, timesteps $t$, heads $h$
    
    \vspace{0.3em}
    \State \textcolor{blue}{\textbf{\textsc{Phase 1: WEST - Threshold Map \& Profile Construction}}}
    \State Initialize Threshold Map $\mathbf{T}_{l,t,h} \leftarrow 1.0$ (Max Value)
    \State Initialize Fidelity Profile $\Phi_{l,t,h} \leftarrow \emptyset$
    
    \For{each layer $l$, timestep $t$, head $h$}
        \For{each calibration sample $x \in \mathcal{D}$}
            \State \textit{// 1. Block-wise Aggregation}
            \State Compute Attn Weights $\mathbf{A} \in \mathbb{R}^{N \times N}$
            \State Aggregate to Block Matrix $\mathbf{S} \in \mathbb{R}^{N_B \times N_B}$ (Eq.~1)
            
            \State \textit{// 2. Local Threshold Discovery}
            \State Sort $\mathbf{S}$ and compute cumulative sum $\mathbf{C}$
            \State For each $p \in \mathcal{P}$, find cutoff $\tau$ s.t. $\mathbf{C}[\tau] \approx p$
            \State Construct local map $\mathbf{L}$: $\mathbf{L}^{(u,v)} = \min \{ p \in \mathcal{P} \mid \text{Block }(u,v) \text{ is active at } p \}$
            
            \State \textit{// 3. Topology Union}
            \State $\mathbf{T}_{l,t,h} \leftarrow \min(\mathbf{T}_{l,t,h}, \mathbf{L})$ \Comment{Include block if active in \textit{any} sample}
        \EndFor
        
        \State \textit{// 4. Compute Fidelity Metrics (Accuracy vs. Density)}
        \For{each $p \in \mathcal{P}$}
            \State Measure approximation error $\epsilon_p$ using Block Mask $\mathbf{M}_{p} = \mathbb{I}(\mathbf{T}_{l,t,h} \le p)$
            \State Record profile: $\Phi_{l,t,h}[p] \leftarrow (\text{Density}_p, \text{Accuracy}_p = 1-\epsilon_p)$
        \EndFor
    \EndFor
    
    \vspace{0.5em}
    \State \textcolor{blue}{\textbf{\textsc{Phase 2: FAST - Mask Instantiation}}}
    \For{each layer $l$, timestep $t$, head $h$}
        \State \textit{// 1. Determine Head-Specific Threshold $p^*$}
        \State Compute Fidelity Threshold $\tau_{acc} = f(l, t; \gamma, \beta)$ \Comment{Spatio-temporal surface}
        \State Find optimal $p^* \in \mathcal{P}$ from $\Phi_{l,t,h}$ such that:
        \State \quad $\text{Accuracy}_{p^*} \ge \tau_{acc}$ AND $\text{Density}_{p^*} \approx \text{LocalTarget}$
        
        \State \textit{// 2. Generate Static Block Mask}
        \State $\mathbf{M}_{l,t,h} = \mathbb{I}(\mathbf{T}_{l,t,h} \le p^*)$ \Comment{Result is $N_B \times N_B$ block mask}
    \EndFor
    
    \State \textbf{Return} $\mathcal{M}$ \Comment{Final block masks ready for kernel execution}
\end{algorithmic}
\end{algorithm}

\section{Detailed Ablation Data}
\label{app:ablation_table}

In this section, we provide the numerical results of the ablation study regarding the Fidelity-Aware Sensitivity Tuning (FAST) module.

\begin{table}[h]
    \centering
    \caption{\textbf{Impact of Adaptive Modulation.} FAST significantly outperforms static uniform sparsity. Notably, our method achieves higher PSNR without any dense warm-up, validating that fidelity-guided modulation effectively redistributes computation to where it matters most.}
    \label{tab:ablation_sparsity}
    \resizebox{0.5\linewidth}{!}{
        \begin{tabular}{l cccc}
            \toprule
            \multirow{2}{*}{\textbf{Method}} & \multicolumn{4}{c}{\textbf{Global Density (\%)}} \\ 
            \cmidrule(lr){2-5} 
             & \textbf{45} & \textbf{50} & \textbf{60} & \textbf{74} \\
            \midrule
            Static Uniform (w/ Warm-up) & 17.83 & 20.50 & 24.39 & 25.78 \\
            \textbf{ScalingAttention (Ours)} & \textbf{23.43} & \textbf{24.13} & \textbf{25.41} & \textbf{27.19} \\
            \bottomrule
        \end{tabular}
    }
\end{table}

\section{Detailed Kernel Overhead Analysis}
\begin{figure}[h]
    \centering
    \includegraphics[width=0.5\linewidth]{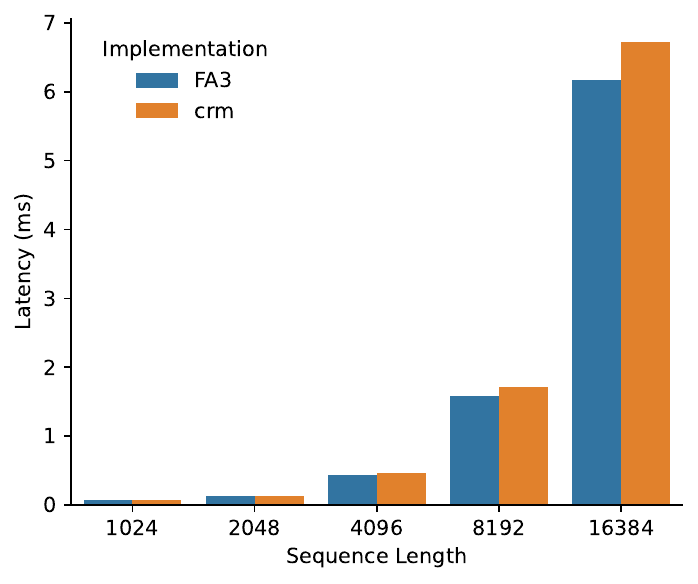}
    \caption{\textbf{Dense Kernel Overhead on Short Sequences.} We compare the absolute latency (ms) of \textit{crm kernel} vs. FA3 at 0\% sparsity. Even on short sequences where kernel launch overheads are typically more pronounced, \textit{crm kernel} maintains a comparable latency profile to FA3, with a maximum overhead of only 9.0\% at $N=16384$.}
    \label{app:fig_short_overhead}
    \vspace{-1.5em}
\end{figure}

To further validate the implementation quality of our block-sparse kernel, we analyze the pure runtime overhead of \texttt{\textit{crm kernel}} compared to FlashAttention-3 in a fully dense setting (0\% sparsity) on shorter sequences in Figure~\ref{app:fig_short_overhead}.

\label{app:sensitivity}

\section{Detailed Sensitivity Analysis}
\label{app:sample_ratio}
As discussed in Section~\ref{sec:sesitivity} of the main paper, we evaluate the consistency
between the estimated Fidelity Score (using sparse sampling) and the ground-truth score
(calculated on the full attention map).

Table~\ref{tab:sample_size_ablation_appendix} demonstrates that even with an extremely low
sampling rate, the estimation remains highly accurate.

\begin{table}[h]
    \centering
    \renewcommand{\arraystretch}{1.1}
    \setlength{\tabcolsep}{4pt}
    
    \caption{\textbf{Robustness of Stratified Sampling.} 
    We report the Consistency across varying sampling rates ($N$). 
    The approximation reaches near-perfect accuracy with $N=16$.}
    \label{tab:sample_size_ablation_appendix}
    
    \begin{tabular}{c c c}
        \toprule
        \textbf{Sampled Rows} & \multicolumn{2}{c}{\textbf{Consistency Metric}} \\
        \cmidrule(lr){2-3}
        ($N$) & \textbf{Mean} & \textbf{Worst-case} \\
        \midrule
        2  & 0.999 & 0.980 \\
        4  & 1.000 & 0.985 \\
        8  & 1.000 & 0.983 \\
        16 & \textbf{1.000} & \textbf{0.999} \\
        32 & 1.000 & 0.999 \\
        64 & 1.000 & 1.000 \\
        \bottomrule
    \end{tabular}
\end{table}

\section{CRM Kernel Pseudocode}
\label{app:crm_code}

\label{app:overhead_analysis}

Algorithm~\ref{alg:crm_fwd} details the forward iterator for the \emph{Compressed Row Mask (CRM)} format. This lightweight iterator enables efficient traversal of sparse attention matrices by skipping empty segments and rapidly enumerating active KV blocks using hardware bitwise primitives.

\begin{algorithm}[h]
\caption{CRM Forward Iterator}
\label{alg:crm_fwd}
\begin{algorithmic}[1]
\small
\State {\bfseries Input:} Row Pointer $row\_ptr$ (start of compressed mask for current query), Size $N$ (number of \texttt{uint32} segments)
\State {\bfseries State:} Struct $S \{ ptr, end, val, base \}$

\vspace{0.5em}
\State \textcolor{blue}{\textbf{Function: Initialize Iterator}}
\Function{Init}{$row\_ptr, N$}
    \State $S.ptr \gets row\_ptr$
    \State $S.end \gets row\_ptr + N$
    \State $S.val \gets 0$
    \State $S.base \gets -32$
    \State \Call{Advance}{$S$} \Comment{Prefetch first non-zero segment}
    \State \Return $S$
\EndFunction

\vspace{0.5em}
\State \textcolor{blue}{\textbf{Function: Skip Empty Segments}}
\Function{Advance}{$S$}
    \While{$S.val = 0$ \textbf{and} $S.ptr \neq S.end$}
        \State $S.val \gets \text{Load}(S.ptr)$ \Comment{Global memory load}
        \State $S.ptr \gets S.ptr + 1$
        \State $S.base \gets S.base + 32$
    \EndWhile
\EndFunction

\vspace{0.5em}
\State \textcolor{blue}{\textbf{Function: Consume Current Active Block}}
\Function{Next}{$S$}
    \State $S.val \gets S.val \ \&\ (S.val - 1)$ \Comment{Clear lowest set bit}
    \If{$S.val = 0$}
        \State \Call{Advance}{$S$} \Comment{Move to next non-zero segment}
    \EndIf
\EndFunction

\vspace{0.5em}
\State \textcolor{blue}{\textbf{Function: Get Global KV Block Index}}
\Function{Index}{$S$}
    \If{$S.val = 0$}
        \State \Return $\bot$ \Comment{End-of-row sentinel}
    \EndIf
    \State $bit \gets \text{CTZ}(S.val)$ \Comment{Count trailing zeros}
    \State \Return $S.base + bit$
\EndFunction

\end{algorithmic}
\end{algorithm}

\section{Additional Threshold Map Visualizations}
\label{app:threshold_maps}

\begin{figure*}[t]
    \centering
    \includegraphics[width= 0.98\linewidth]{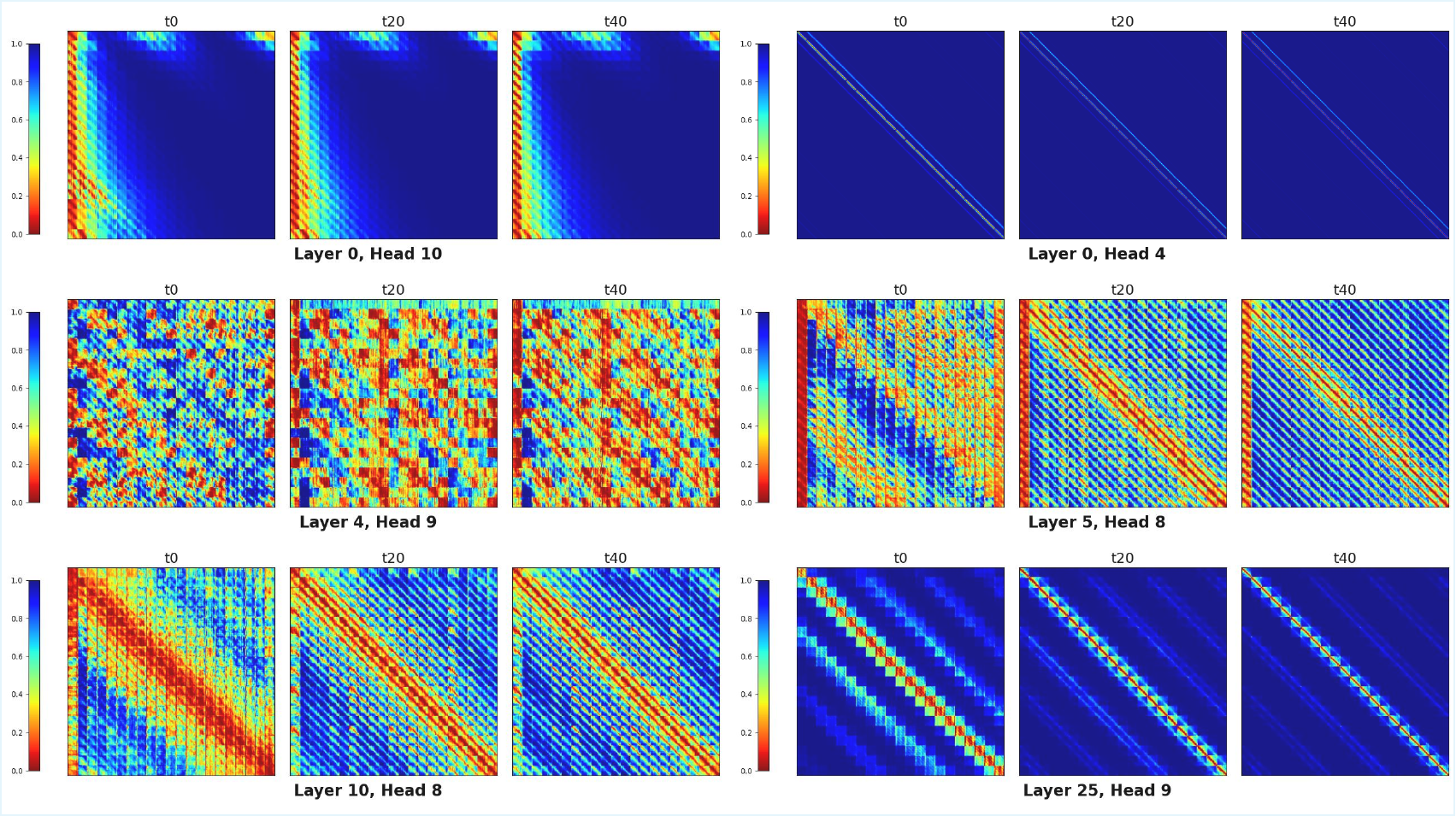}
    \caption{\textbf{Threshold Map Gallery for WEST.}
    We visualize representative attention structures across layers/heads and diffusion timesteps (e.g., $t{=}0,20,40$).
    Despite diverse per-prompt activations, the stable support envelope captured by WEST remains consistent, supporting the existence of an intrinsic, weight-encoded sparse topology.}
    \label{fig:thresholdmap_gallery}
\end{figure*}

In this section, we present extended visualizations of the
\textbf{Threshold Maps} extracted by the WEST module.
As discussed in Section~\ref{motivation} (Observation~1), these maps capture the
intrinsic sparse topology encoded in the pre-trained weights.
Figure~\ref{fig:thresholdmap_gallery} showcases this topology across
a diverse set of attention heads, layers, and diffusion timesteps.
Despite variations in input prompts, the high-mass attention regions
of each head consistently lie within a stable and well-defined structure,
further corroborating our empirical findings.





\begin{figure*}[t]
    \centering
    \includegraphics[width=\linewidth]{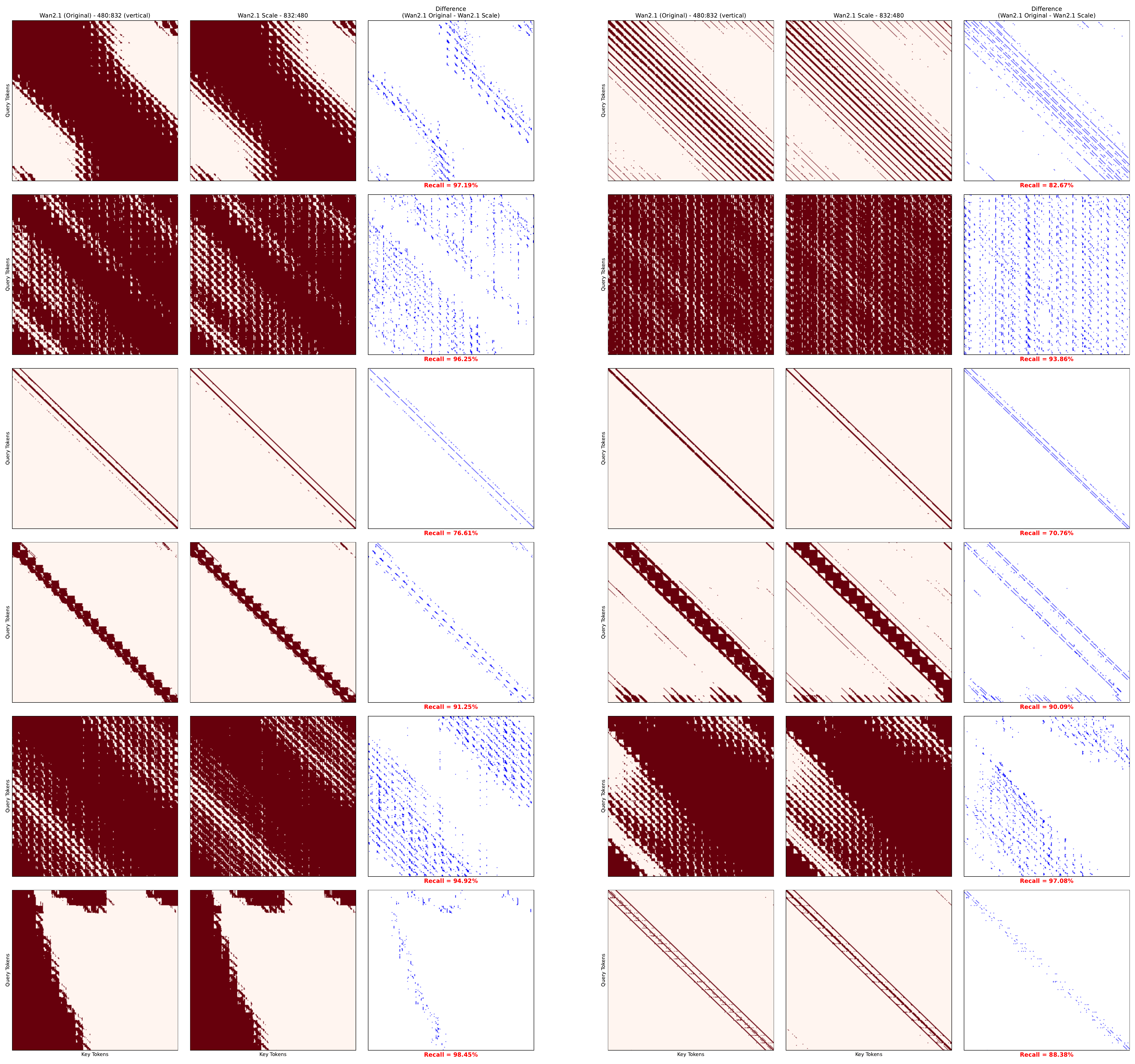}
    \caption{\textbf{Appendix visualization of aspect-ratio robustness.} For the same prompt under different aspect-ratio settings, the generated videos exhibit substantial semantic-layout differences, yet the recovered sparse attention support remains nearly unchanged. This provides an additional view that the Intrinsic Sparse Topology is primarily weight-encoded rather than input-specific.}
    \label{fig:appendix_aspect_ratio}
\end{figure*}

\begin{figure*}[t]
    \centering
    \includegraphics[width=0.9\linewidth]{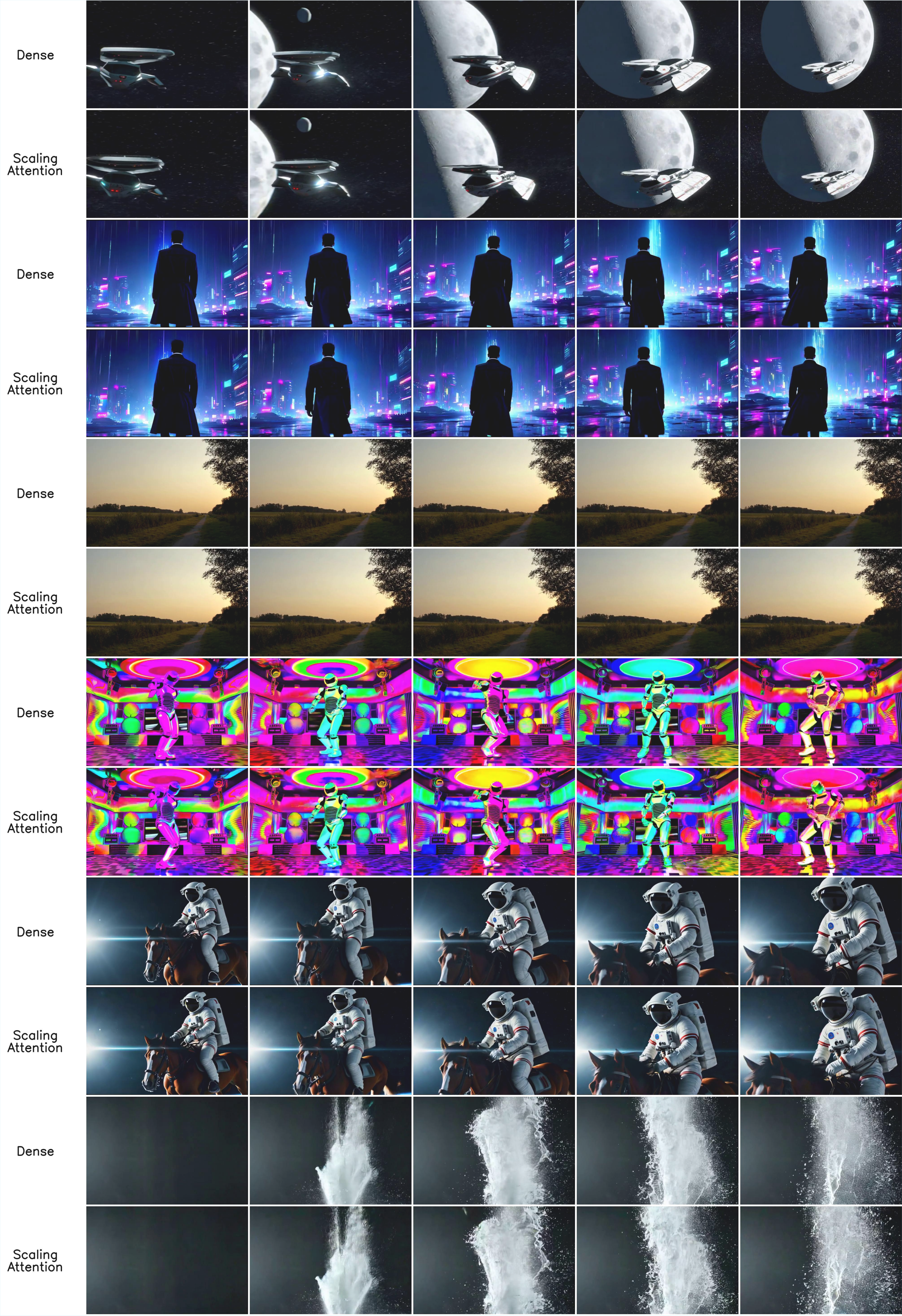}
    \caption{\textbf{Qualitative comparison on HunyuanVideo.} Side-by-side frames from dense attention and ScalingAttention at global density of 55\%.}
    \label{fig:hunyuan_visual_comparison}
\end{figure*}


\end{document}